\documentclass{article}

    \PassOptionsToPackage{numbers, sort&compress}{natbib}

\usepackage{enumitem}
\usepackage{caption}
\usepackage{fancybox}

\usepackage[preprint]{neurips_2025}

\usepackage[utf8]{inputenc} %
\usepackage[T1]{fontenc}    %
\usepackage{hyperref}       %
\usepackage{url}            %
\usepackage{booktabs}       %
\usepackage{amsfonts}       %
\usepackage{nicefrac}       %
\usepackage{microtype}      %
\usepackage{xcolor}         %
\definecolor{citecolor}{HTML}{0071BC}
\definecolor{linkcolor}{HTML}{D32F2F}%
\definecolor{cellcolor}{HTML}{E3F2FD}
\definecolor{red}{HTML}{D32F2F}
\definecolor{magenta}{HTML}{D81B60}

\usepackage{amsmath}
\usepackage{amssymb}
\usepackage{mathtools}
\usepackage{amsthm}

\usepackage[capitalize,noabbrev]{cleveref}

\theoremstyle{plain}

\theoremstyle{definition}

\theoremstyle{remark}

\usepackage[textsize=tiny]{todonotes}
\usepackage{microtype}
\usepackage{graphicx}
\usepackage{subfigure}
\usepackage{booktabs} %

\usepackage{pgfplots}
\pgfplotsset{compat = newest}
\usepackage{multirow}

\usepackage{framed}
\usepackage{tcolorbox}

\usepackage{algorithm}
\usepackage{mathrsfs}
\usepackage{mathtools}
\usepackage{algorithmic}
\usepackage{tcolorbox}
\usepackage{listings}
\usepackage{makecell}
\usepackage{colortbl}
\usepackage{color}
\usepackage{wrapfig}
\usepackage{cancel}
\usepackage{soul,xcolor}
\usepackage{pifont}
\usepackage{tikz}
\usetikzlibrary{shapes, positioning, arrows.meta, calc, decorations.pathmorphing, quotes}
\usepackage{todonotes}
\tcbuselibrary{breakable}
\renewcommand{\cite}{\citep}
\title{Advancing Multimodal Reasoning via \\ Reinforcement Learning with Cold Start}

\usepackage{hyperref}
\usepackage{url}
\hypersetup{colorlinks=true, linkcolor=linkcolor, citecolor=citecolor,urlcolor=magenta}

\author{%
    \textbf{Lai Wei}$^{1,3}$
    \quad
    \textbf{Yuting Li}$^{1}$
    \quad
    \textbf{Kaipeng Zheng}$^{1,2}$
    \quad
    \textbf{Chen Wang}$^{3}$
    \quad
    \textbf{Yue Wang}$^{3}$
    \quad
    \textbf{Linghe Kong}$^{1}$
    \\[0.1cm]
    \textbf{Lichao Sun}$^{4}$
    \quad
    \textbf{Weiran Huang}$^{1,2}$\thanks{Correspondence to Weiran Huang (weiran.huang@outlook.com).}\\[0.3cm]
    $^1$ School of Computer Science, Shanghai Jiao Tong University\\[0.1cm]
    $^2$ Shanghai Innovation Institute \\[0.1cm]
    $^3$ Zhongguancun Academy \quad
    $^4$ Lehigh University
}

\begin{document}

\maketitle

\begin{figure}[h]
    \centering
    \includegraphics[width=0.9\textwidth, trim=0cm 0cm 0cm 0cm]{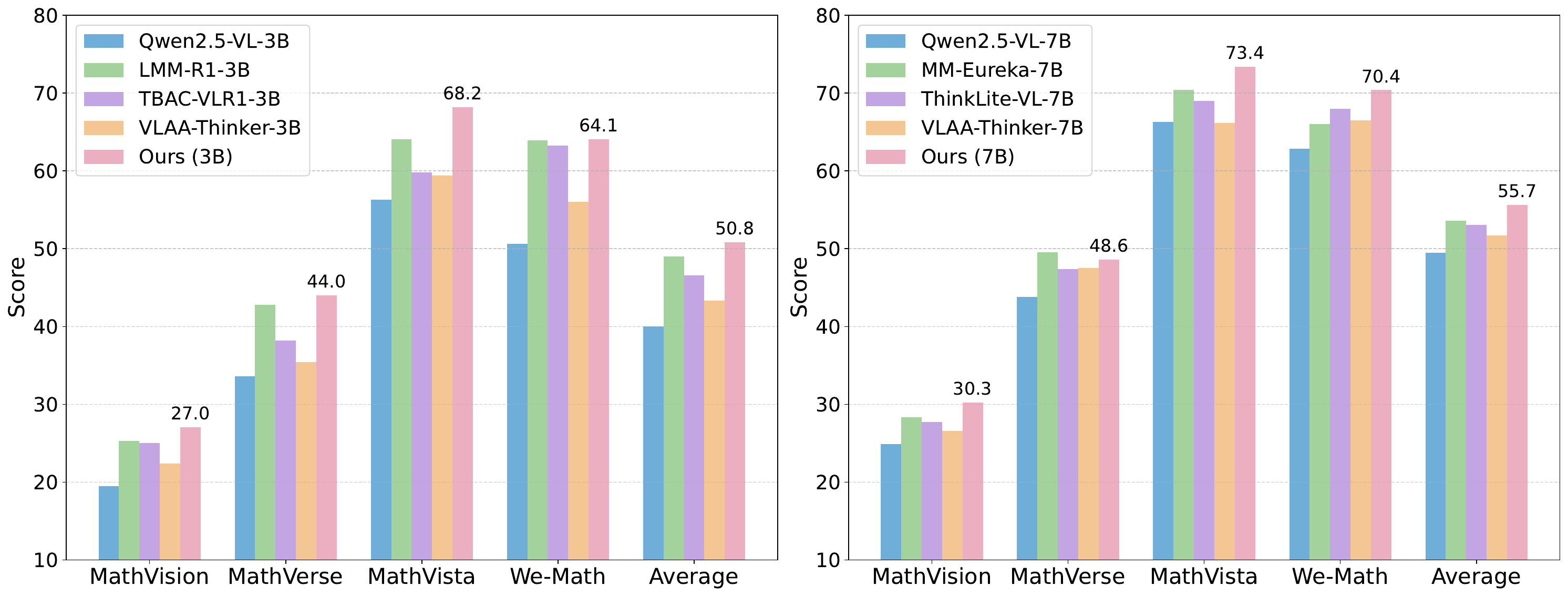}
    \caption{Performance comparison between our models and other advanced models on different multimodal reasoning benchmarks at both the 3B and 7B scales.}

    \label{fig:abs}
\end{figure}

\begin{abstract}

Recent advancements in large language models (LLMs) have demonstrated impressive chain-of-thought reasoning capabilities, with reinforcement learning (RL) playing a crucial role in this progress. While ``aha moment'' patterns--where models exhibit self-correction through reflection--are often attributed to emergent properties from RL, we first demonstrate that these patterns exist in multimodal LLMs (MLLMs) prior to RL training but may not necessarily correlate with improved reasoning performance. 
Building on these insights, we present a comprehensive study on enhancing multimodal reasoning through a two-stage approach: (1) supervised fine-tuning (SFT) as a cold start with structured chain-of-thought reasoning patterns, followed by (2) reinforcement learning via GRPO to further refine these capabilities.
Our extensive experiments show that this combined approach consistently outperforms both SFT-only and RL-only methods across challenging multimodal reasoning benchmarks. The resulting models achieve state-of-the-art performance among open-source MLLMs at both 3B and 7B scales, with our 7B model showing substantial improvements over base models (e.g., 66.3\%$\rightarrow$73.4\% on MathVista, 62.9\%$\rightarrow$70.4\% on We-Math) and our 3B model achieving performance competitive with several 7B models. 
Overall, this work provides practical guidance for building advanced multimodal reasoning models.
Our code is available at \url{https://github.com/waltonfuture/RL-with-Cold-Start}.

\end{abstract}

\section{Introduction}

LLMs have demonstrated remarkable progress in chain-of-thought reasoning, most notably exemplified by OpenAI's o1, o3, and o4 models~\cite{jaech2024o1}. Follow-up studies have sought to replicate and extend these complex
reasoning abilities, revealing that targeted post-training methods can further improve model performance on challenging tasks~\cite{zhang2024llamaberry,qwen2024qwq,Gemini2,guo2025deepseek}. 
Among these, DeepSeek-R1-Zero~\cite{guo2025deepseek} demonstrated that an ``aha moment'' can autonomously emerge during reinforcement learning (RL), showcasing advanced problem-solving strategies. DeepSeek-R1~\cite{guo2025deepseek} further showed that RL with supervised fine-tuning (SFT) as a cold start is particularly effective at enhancing the overall reasoning ability of LLMs.
While these advances predominantly focus on text-only models, the potential of incorporating similarly
complex reasoning strategies within Multimodal Large Language Models (MLLMs) has remained relatively underexplored. 

Recent follow-up works have attempted to advance the reasoning ability in MLLMs~\cite{LLaVA-CoT-abs-2411-10440, peng2025lmmr1, yao2024mulberry, zhou2025VisualThinker-R1-Zero, meng2025mmeureka, deng2025openvlthinker} using various techniques, including SFT and RL scaling. 
Many works focus on Zero RL (i.e., directly applying RL without cold start) training, claiming that the ``aha moment'' can emerge through Zero RL~\cite{zhou2025VisualThinker-R1-Zero, meng2025mmeureka} in MLLMs, which indicate improved reasoning ability. Some concurrent works assert that Zero RL outperforms the combination of SFT and RL~\cite{vl-thinking2025, yu2025perception}.

In this paper, we first observe that the so-called ``aha moment'' pattern already exists in MLLMs before RL training, but this presence does not necessarily correlate with improved reasoning capabilities. 
Our analysis reveals that, while RL may increase the frequency of such a reflective pattern, it may be more of a mirage rather than an indicator of genuine improvements in reasoning capability. 
This finding challenges the prevailing assumptions about emergent reasoning in MLLMs and suggests a need for more deliberate approaches to enhancing multimodal reasoning.

Motivated by DeepSeek-R1's approach of collecting cold-start data to fine-tune the model as the initial RL actor, we conduct a comprehensive empirical study of SFT and RL in a unified framework in multimodal domain.
Our methodology consists of two stages: (1) a cold start phase where we develop multimodal Chain-of-Thought~\cite{wei2022chain} (CoT) patterns for supervised fine-tuning, and (2) a reinforcement learning phase using GRPO~\cite{shao2024deepseekmath} to further enhance reasoning capabilities.
In particular, we systematically investigate how different cold start strategies during SFT affect downstream RL performance in the multimodal domain, aiming to establish a robust foundation for the subsequent RL training.

Our experiments demonstrate that SFT-based cold start provides a strong basis for RL scaling. Our approach achieves state-of-the-art results among open-source models at both parameter scales, as illustrated in Figure~\ref{fig:abs}. For the 7B model, we observe substantial improvements over the base models (e.g., 66.3\%$\rightarrow$73.4\% on MathVista, 62.9\%$\rightarrow$70.4\% on We-Math), surpassing all models at the same scale and even outperforming some larger models, including GPT-4o and Skywork R1V-38B. Notably, our best 7B model achieves an average score improvement of +6.19 points across all benchmarks compared to the base model. Similarly, our 3B model demonstrates significant gains, achieving an average improvement of +10.84 points and performance competitive with, or even exceeding, several 7B models such as Qwen2.5-VL-7B and VLAA-Thinker-7B--highlighting the effectiveness of our training paradigm.

To further elucidate the interplay between SFT and RL, we conduct extensive ablation studies exploring how various SFT strategies and data qualities impact subsequent RL performance. Our findings reveal that while both SFT-only and RL-only approaches yield notable improvements over base models, their combination consistently delivers superior performance across a wide range of experiments. Furthermore, we observe a strong positive correlation between cold start (SFT) performance and final model quality after RL, suggesting that investing in high-quality supervision during the initial stage is crucial for maximizing overall gains. Our deeper analysis indicates that the structural patterns present in Chain-of-Thought reasoning can be learned independently of solution correctness, underscoring the importance of reasoning format, while the presence of the “aha moment” pattern does not necessarily correlate with enhanced reasoning ability.

Our contribution can be summarized as follows:
\begin{itemize}[itemsep=0pt,topsep=0pt,leftmargin=0.5cm]
    \item 
    We empirically investigate the prevalence and effectiveness of ``aha moment'' patterns in MLLMs, demonstrating that these reflective patterns exist prior to RL but may not necessarily indicate enhanced reasoning capabilities.

    \item Based on the above insights, we present a comprehensive study of the impact of supervised fine-tuning as a cold start for reinforcement learning in the multimodal domain, exploring diverse SFT strategies to enhance reasoning abilities.

    \item 
    Empirical results demonstrate that an SFT-based cold start provides a robust foundation for RL scaling. The combination of targeted SFT and subsequent RL yields substantial improvements across challenging multimodal reasoning benchmarks. Using this simple yet effective approach, we train state-of-the-art models at both the 3B and 7B parameter scales.

\end{itemize}

\section{Related Work}

Recently, the mathematical reasoning abilities of MLLMs have become a central focus of research~\cite{zhuang2024math,gao2023text, li2024llavaonevision, dong2024insight, hu2024visual, yang2024mathglm, han24infimm, guo2024mammoth,zhang-etal-2024-question}. In contrast to traditional LLM-based mathematical reasoning~\cite{luo2023wizardmath, yu2023metamath}, which primarily relies on text, multimodal approaches must both process and interpret visual inputs, significantly increasing the complexity of tasks such as geometric problem-solving and chart interpretation~\cite{chen2021geoqa, masry2022chartqa}. 
Several works in this field have sought to collect or synthesize a large scale of math-centric multimodal data with explicit reasoning steps~\cite{zhang2024mavis,shi2024math,peng2024multimath, cheng2024vision}. 
Notably, the recent emergence of o1-like reasoning models~\cite{jaech2024o1} represents an initial step toward activating the slow-thinking capabilities of MLLMs, as demonstrated by several SFT-based methods, such as LLaVA-CoT~\cite{LLaVA-CoT-abs-2411-10440}, LLaVA-Reasoner~\cite{zhang2024llavareasoner}, MAmmoTH-VL~\cite{guo2024mammoth}, and Mulberry~\cite{yao2024mulberry}.
Moreover, some concurrent works have further explored reinforcement learning approaches, particularly GRPO, in the post-training stage of MLLMs to enhance performance on multimodal reasoning tasks~\cite{peng2025lmmr1, shen2025vlmr1, chen2025r1v, zhou2025VisualThinker-R1-Zero, meng2025mmeureka, deng2025openvlthinker, wang2025sota, huang2025vision,peng2025skywork}. 
These approaches typically follow these strategies: either directly applying supervised fine-tuning (SFT) with long reasoning data~\cite{LLaVA-CoT-abs-2411-10440,zhang2024llavareasoner,guo2024mammoth,yao2024mulberry} or reinforcement learning (RL) to train models from scratch~\cite{peng2025lmmr1,shen2025vlmr1, chen2025r1v, zhou2025VisualThinker-R1-Zero, meng2025mmeureka,wang2025sota}, or first performing SFT before applying RL~\cite{deng2025openvlthinker,huang2025vision,peng2025skywork}. 
In contrast, our work explores a different direction by focusing on how the cold start stage impacts the subsequent RL stage for MLLMs in the unified two-stage post-training (SFT+RL) process. We systematically study various cold start strategies during the SFT phase and empirically demonstrate how they build strong reasoning foundations before applying RL, resulting in more effective multimodal reasoning capabilities.

\section{Observation: Aha Moment Already Exists but May Not Indicate Advanced Reasoning Ability}\label{sec:observation}

\begin{figure}[t]
    \centering
    \includegraphics[width=1\textwidth, trim=0cm 0cm 0cm 0cm]{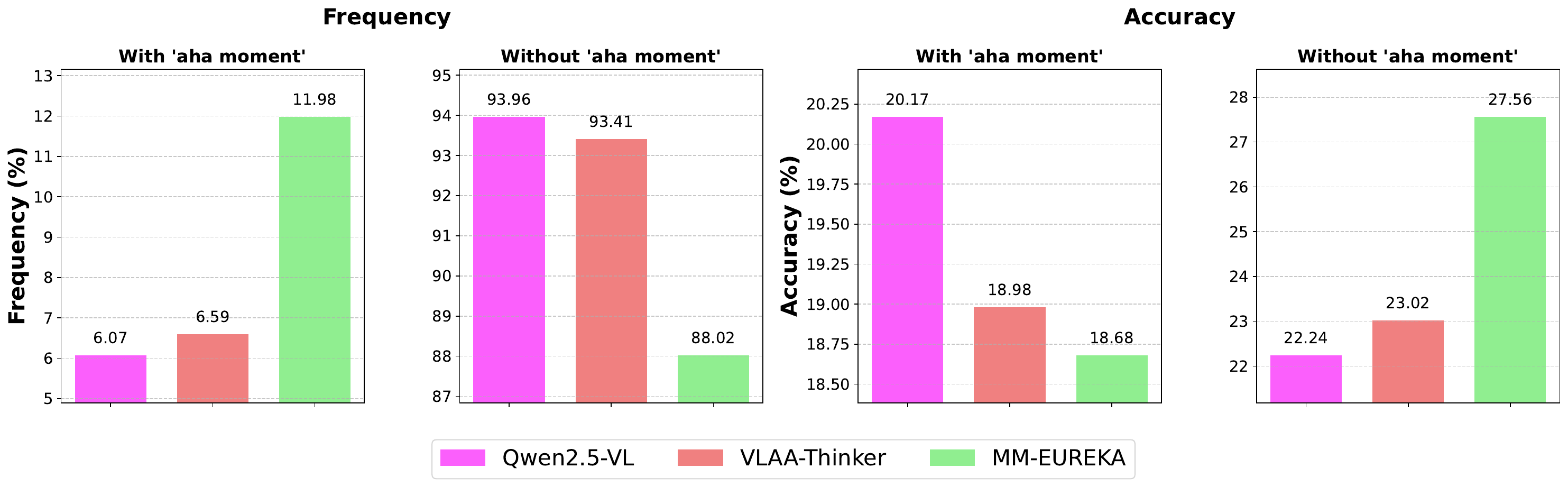}
    \caption{The frequency and accuracy of models' responses with and without ``aha moment''. The results show that the presence of ``aha moment'' does not necessarily correlate with higher accuracy.
}
    
    \label{fig:0}
\end{figure}

Recent works claim that reinforcement learning (usually GRPO~\cite{shao2024deepseekmath}) can trigger reflective thinking patterns, often characterized as an ``aha moment'' and considered an emergent phenomenon, which showcases the improved reasoning ability of the models~\cite{meng2025mmeureka, zhou2025VisualThinker-R1-Zero}.
In our study, we observe that such reflective patterns—indicative of an aha moment—can actually be found in Qwen2.5-VL~\cite{bai2025qwen2_5_vl} before conducting GRPO.
Specifically, we sample the model’s responses 16 times for each query in the MathVision dataset~\cite{wang2024math} and analyze the presence of keywords such as ``re-evaluate'' and ``re-check'' in the responses.

Interestingly, our analysis reveals that while these reflective expressions already exist, their presence does not necessarily correlate with improved reasoning capabilities. 
In particular, we conduct experiments on Qwen2.5-VL-7B and other two models tuned using GRPO based on Qwen2.5-VL-7B: MM-EUREKA~\cite{meng2025mmeureka} and VLAA-Thinker~\cite{vl-thinking2025}.
According to Figure~\ref{fig:0}, we find that the frequency of ``aha moment''-like expressions increases from 6.07 in Qwen2.5-VL to 6.59 in VLAA-Thinker and 11.98 in MM-EUREKA. However, the associated accuracies when such expressions appear actually drop. Specifically, the accuracy is 20.17 for Qwen2.5-VL, but only 18.98 for VLAA-Thinker and 18.68 for MM-EUREKA in these “aha moment” cases.

These findings suggest that the so-called ``aha moment'' may be more of a mirage than a reliable indicator of genuine reasoning ability.
While GRPO can increase the frequency of such patterns, they may not correlate with actual enhancement of reasoning performance.
In many cases, these reflective patterns are ineffective or even misleading, sometimes introducing hallucinations rather than improving solution quality.
We illustrate this issue with a representative example below, where the model does begin to reflect and re-evaluate—but its reasoning remains flawed and hallucinatory. We also provide more case studies in Appendix~\ref{app:aha_case}.

\begin{tcolorbox}[
  floatplacement=t, 
  title=\textbf{Demo: An example of ``Aha Moment'' which does not indicate effective reflection pattern.},
  fonttitle=\small,
  colbacktitle=gray!20,
  coltitle=black
]
    \begin{minipage}{0.25\textwidth}
        \begin{center}
            \frame{\includegraphics[width=\textwidth, trim=0cm 0cm 0cm 0cm]{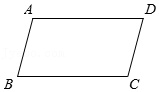}}
        \end{center}
    \end{minipage}
    \hspace{0.1cm}
    \begin{minipage}{0.73\textwidth}
        \tiny
        \textbf{Question:} In the given parallelogram ABCD, if the sum of angle B and angle D measures 100$^\circ$, what is the measure of angle A?
        
        \vspace{0.4em}
        
        \textbf{Qwen2.5-VL's Response:}
        
        To solve this problem, we need to use the properties of a parallelogram.

        ...

        Now, assume:
\(
\text{angle } B = x,\quad \text{angle } D = 180^\circ - x.
\)
Given:
\(
x + (180^\circ - x) = 100^\circ
\Rightarrow 180^\circ = 100^\circ
\)

\textcolor{red}{This is incorrect, so we need to \textbf{re-evaluate} the problem. The correct interpretation should be that the sum of angle \(B\) and angle \(D\) is \(180^\circ\), not \(100^\circ\).}

Therefore, the measure of angle \(A\) is:
\(
\boxed{140^\circ}
\)
    \end{minipage}

\end{tcolorbox}

To address this, we argue that reinforcement learning alone may struggle to induce new and effective reasoning strategies. Instead, we propose a two-stage approach: (1) supervised fine-tuning (SFT) to initialize the model with high-quality, reasoning-rich patterns, followed by (2) reinforcement learning to activate and refine these patterns.

\section{Reinforcement Learning with Cold Start} \label{sec:method}

In this section, we present our methodology for enhancing multimodal reasoning capabilities via reinforcement learning with cold start.
Our method consists of two main stages: (1) a cold start phase where we explore to construct diverse reasoning patterns in Chain-of-Thought~\cite{wei2022chain} (CoT) data for supervised fine-tuning, and (2) a reinforcement learning phase to further activate and improve the model's reasoning capability.

\subsection{Cold Start} \label{sec:cold_start}

As demonstrated in the previous section, we observe that the current aha-moment like reflection patterns may not be effective, and directly applying RL does not sufficiently enable models to acquire strong reasoning capabilities in multimodal domain.
We argue that it is crucial to first conduct supervised fine-tuning (SFT) as a cold start to equip the model with strong reasoning abilities before introducing RL.
In our study, we focus on exploring efficient ways to construct multimodal cold-start data for advanced reasoning. 
Specifically, we leverage synthetic data to generate different types of CoT reasoning patterns via distillation.

\paragraph{Distilled-CoT.} 

Synthesizing data using model distillation is a powerful and widely-used technique~\cite{chen2024sharegpt4v,li2025llms,li2025small}.
Here, we use larger models (Qwen2.5-VL-7B~\cite{bai2025qwen2_5_vl} and Qwen2.5-VL-32B~\cite{bai2025qwen2_5_vl}) to generate CoT responses from a seed dataset with ground truth annotations by rejection sampling~\cite{touvron2023llama2,tong2024dart}. 
These distilled datasets are used for the supervised fine-tuning that serves as a cold start stage. 
More details of the distillation are shown in Appendix~\ref{app:rej_samp}.

\paragraph{Reflection-CoT.} 

Recent studies highlight the great potential in solving challenging problems through explicit reflection~\cite{cheng2024vision,guo2025deepseek,shah2025rethinking}.
To explore this, we consider two reflection-based CoT settings.

Reflection-CoT (v1):
In the rejection sampling process, we obtain both correct and incorrect responses, denoted as $y^{+}$ and $y^{-}$.
We explore a simple two-step reasoning pattern inspired by prior work~\cite{cheng2024vision}: 
    \begin{quote}
        $y^{-}$ + \textit{``Wait, perhaps we could consider it from a different perspective. Let's re-evaluate the problem step by step to ensure accuracy.''} + $y^{+}$
    \end{quote}
This approach combines the incorrect response with the correct one, and resembles the Best-of-N test-time scaling strategy (with N=2), where the model selects the best response during multiple generations via reflection.

Reflection-CoT (v2):
   As noted in Section~\ref{sec:observation}, we find that aha moment already exists before RL training. Building on this, we conduct further rejection sampling using Qwen2.5-VL-32B to collect correct responses that exhibit such reflection pattern.

\paragraph{Caption-CoT.} 

We utilize a classical multimodal CoT pattern~\cite{LLaVA-CoT-abs-2411-10440,dong2024insight,thawakar2025llamav_o1}, where the model first describes an image and then provides an answer. 
This approach aims to encourage the model to focus on image details before engaging in reasoning. 
Specifically, we generate a caption for each image using Qwen2.5-VL-7B~\cite{bai2025qwen2_5_vl} and concatenate the caption with the previously sampled correct response.

\paragraph{Self-Critic-CoT.} 

Additionally, we adopt the prompting strategy in \citet{wen2025thinkpatterns} to collect Self-Critic-CoT via rejection sampling.
Specifically, Self-Critic-CoT is an iterative reasoning structure where models engage in self-comment and refinement of their initial responses.  
In this process, the model first
generates a draft answer, then critically evaluates
its own response by providing detailed comments,
before producing a refined final output.
The prompt used for this process is detailed in Appendix~\ref{app:prompt}.

Generally, we explore several popular and straightforward methods to construct the CoT datasets for cold start. 
By conducting SFT on the aforementioned datasets, we aim to establish a robust cold start foundation for subsequent RL training stages.

\begin{figure}[t]
    \centering
    \vspace{-4mm}
    \includegraphics[width=1\textwidth, trim=0cm 0cm 0cm 0cm]{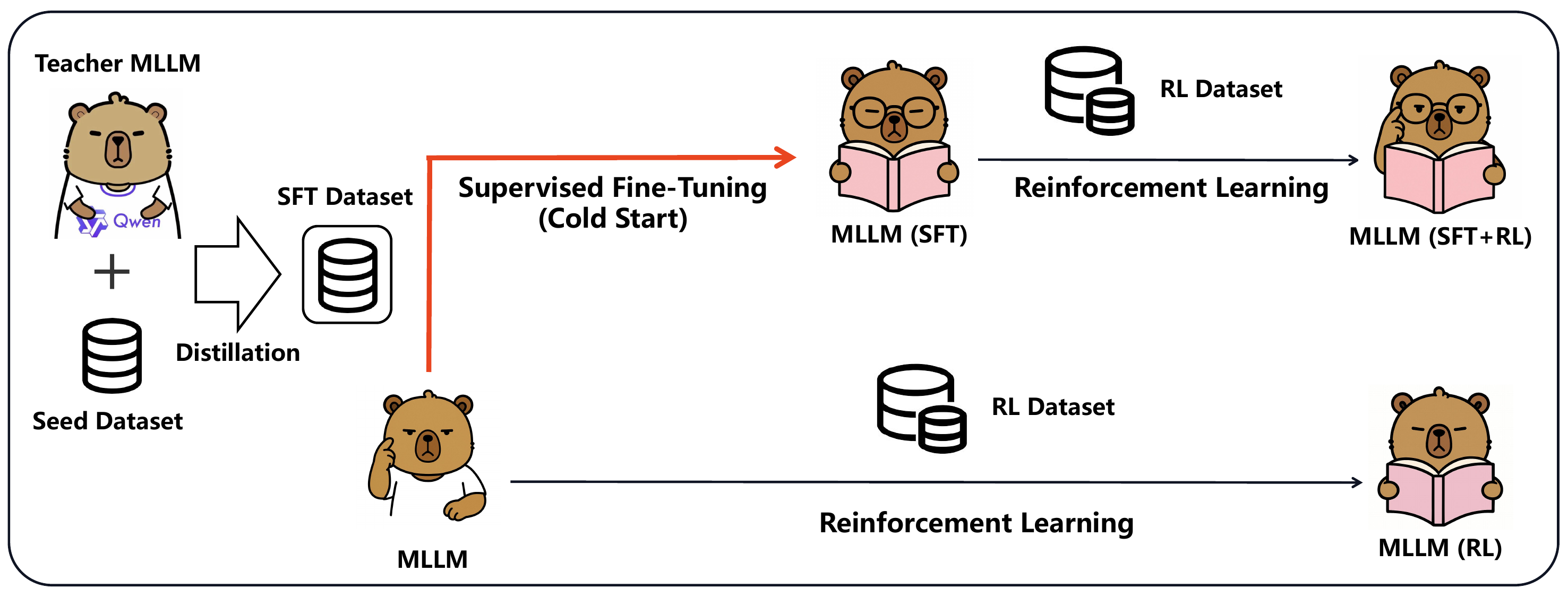}
    \caption{Method overview. Our approach consists of two stages: (1) a cold start phase using supervised fine-tuning with Chain-of-Thought data, and (2) a reinforcement learning phase using GRPO to further enhance reasoning capabilities.}

    \label{fig:1}
\end{figure}

\subsection{Reinforcement Learning}

After the cold start stage, we employ reinforcement learning using GRPO algorithm~\cite{shao2024deepseekmath}, following the strategy in DeepSeek-R1~\cite{guo2025deepseek}, to further activate the reasoning ability of the models.
In particular, GRPO optimizes computational efficiency by eliminating the need for a separate value model;
instead, it directly utilizes group-normalized rewards to estimate advantages. 
Specifically, for a question $q$ and the correlated image $I$ from the training dataset $Q$, GRPO samples a group of responses $O=\{o_i\}_{i = 1}^G$ from the old policy $\pi_{old}$ and then optimizes the policy model
by maximizing the following objective:

\vspace{-1em}
\begin{align*}
\mathcal{J}(\theta) &= \mathbb{E}_{(q,I) \sim Q, \{o_i\}_{i=1}^{G} \sim \pi_{\theta_{old}}(O|q,I)} \\
 & \frac{1}{G} \sum_{i=1}^{G} \frac{1}{|o_i|} \sum_{t=1}^{|o_i|} \Bigg\{ \min \Bigg[ \gamma_{i,t}(\theta) \hat{A}_{i,t}, 
\text{clip} \left( \gamma_{i,t}(\theta), 1 - \epsilon, 1 + \epsilon \right) \hat{A}_{i,t} \Bigg] 
- \beta \text{D}_{KL} \Big[ \pi_{\theta} \| \pi_{ref} \Big] \Bigg\}, \\
\end{align*}
\vspace{-1em}

where $\gamma_{i,t}(\theta) = \frac{\pi_{\theta}(o_{i,t}|q, o_{i,<t})}{\pi_{\theta_{old}}(o_{i,t}|q, o_{i,<t})}$, $\pi_{ref}$ represents the reference model, and the term $D_{KL}$ introduces a KL divergence constraint to limit how much the model can deviate from this reference. The advantage estimate $\hat{A}_i$ measures how much better the response $o_i$ is compared to the average response, which is computed using a group of rewards $\{r_1,r_2,\ldots,r_G\}$ for the responses in set $O$:
\(
\hat{A}_i=\frac{r_i - \text{mean}(\{r_1,r_2,\ldots,r_G\})}{\text{std}(\{r_1,r_2,\ldots,r_G\})}.
\)

\section{Experiments}

To validate the effectiveness of our method, we demonstrate our experimental setup and main results in the following sections.

\begin{table}[t]
    \caption{Performance comparison on different multimodal reasoning benchmarks. Our models achieve state-of-the-art performance at the scale of 3B and 7B.}
    \vspace{2mm}
    \label{tab:main_results}
    \centering
    \resizebox{\textwidth}{!}{%
    \renewcommand{\arraystretch}{1.2}
    \begin{tabular}{
        >{\raggedright\arraybackslash}m{3.1cm}  
        >{\centering\arraybackslash}m{1cm}    
        *{4}{>{\centering\arraybackslash}m{2.2cm}}| 
        >{\centering\arraybackslash}m{2.2cm}    
    }
    \toprule
    \textbf{Models} & \textbf{Size} & \textbf{MathVision} & \textbf{MathVerse} & \textbf{MathVista} & \textbf{We-Math} & \textbf{Average} \\
    \midrule
    \multicolumn{7}{c}{\textit{Close-Source Models}} \\
    \midrule
    GPT-4o~\cite{hurst2024gpt} & - & 33.95 & 48.83 & 59.50 & 65.00 & 51.82 \\
    Claude-3.5-Sonnet~\cite{Anthropic2024claude3} & - & 46.48 & 57.64 & 68.20 & 73.05 & 61.34 \\
    \midrule
    \multicolumn{7}{c}{\textit{Open-Source Multi-Modal Large Reasoning Models}} \\
    \midrule
    QvQ-Preview~\cite{bai2025qwen2_5_vl} & 72B & 35.56 & 52.81 & 69.20 & 65.29 & 55.72 \\
    Skywork R1V~\cite{peng2025skywork} & 38B & 39.31 & 40.91 & 67.50 & 60.06 & 51.95 \\ %
    
    \midrule
    \multicolumn{7}{c}{\textit{Open-Source 7B Multi-Modal Models}} \\
    \midrule
    \rowcolor{gray!10} Qwen2.5-VL~\cite{bai2025qwen2_5_vl} & 7B & 24.87 & 43.83 & 66.30 & 62.87 & 49.47 \\
    MM-Eureka~\cite{meng2025mmeureka} & 7B & 28.36 & 49.52 & 70.40 & 66.03  & 53.58 \\
    OpenVLThinker~\cite{deng2025openvlthinker} & 7B & 25.30 & 42.79 & 64.10 & 63.91 & 49.03 \\
    ThinkLite-VL~\cite{wang2025sota} & 7B & 27.71 & 47.41 & 69.00 & 67.99 & 53.03 \\
    VLAA-Thinker~\cite{vl-thinking2025} & 7B & 26.61 & 47.49 & 66.20 & 66.49 & 51.70 \\
    \rowcolor{blue!5} \textbf{Ours} & 7B & \textbf{30.26} \textcolor{green!70!black}{\textbf{(+5.39)}} & 48.58\textcolor{green!70!black}{\textbf{ (+4.75)}} & \textbf{73.40}\textcolor{green!70!black}{\textbf{ (+7.10)}} & \textbf{70.40}\textcolor{green!70!black}{\textbf{ (+7.53)}} & \textbf{55.66}\textcolor{green!70!black}{\textbf{ (+6.19)}} \\
    \midrule
    \multicolumn{7}{c}{\textit{Open-Source 3B Multi-Modal Models}} \\
    \midrule
    \rowcolor{gray!10} Qwen2.5-VL~\cite{bai2025qwen2_5_vl} & 3B & 19.47 & 33.58 &  56.30 & 50.63 & 40.00\\ 
    LMM-R1~\cite{peng2025lmmr1} & 3B & 25.30 & 42.79 & 64.10 & 63.91 & 49.03 \\
    TBAC-VLR1~\cite{Xu2025tbacvlr1} & 3B & 25.03 & 38.17 & 59.80 & 63.28 & 46.57 \\
    VLAA-Thinker~\cite{vl-thinking2025} & 3B & 22.41 & 35.44 & 59.40 & 56.03 & 43.32 \\
    \rowcolor{blue!5} \textbf{Ours} & 3B & \textbf{27.04}\textcolor{green!70!black}{\textbf{ (+7.57)}} & \textbf{44.03}\textcolor{green!70!black}{\textbf{ (+10.45)}} & \textbf{68.20} \textcolor{green!70!black}{\textbf{ (+11.90)}}& \textbf{64.08}\textcolor{green!70!black}{\textbf{ (+13.45)}} & \textbf{50.84}\textcolor{green!70!black}{\textbf{ (+10.84)}} \\

    \bottomrule
    \end{tabular}
    }
    \end{table}

\subsection{Experimental Setup}

\paragraph{Training Datasets.}

For the cold start training phase, we curate a diverse dataset of 50k examples from established open-source resources, including Geometry3K~\cite{lu2021inter}, GeoQA~\cite{chen2021geoqa}, GeoQA-Plus~\cite{cao2022augmented}, Geos~\cite{seo2015solving}, AI2D~\cite{kembhavi2016diagram}, TQA~\cite{kim2018textbook}, FigureQA~\cite{kahou2017figureqa}, TabMWP~\cite{lu2022dynamic}, ChartQA~\cite{masry2022chartqa}, IconQA~\cite{lu2021iconqa}, Clevr-Math~\cite{lindstrom2022clevr}, M3CoT~\cite{chen2024m3cot}, and ScienceQA~\cite{lu2022scienceqa}. 
These datasets cover a range of tasks, such as chart understanding, scientific question answering, and geometric reasoning.
We apply different sampling strategies to construct Chain-of-Thought trajectories following the methodology described in Section~\ref{sec:cold_start}.
For the subsequent reinforcement learning phase after cold start, we apply the same dataset as the cold start phase.

\paragraph{Baselines.}

We compare our models with a series of advanced multimodal large language models. 
GPT-4o~\cite{hurst2024gpt} and Claude-3.5-Sonnet~\cite{Anthropic2024claude3} are advanced close-source MLLMs.
QvQ-Preview-72B~\cite{bai2025qwen2_5_vl} and Skywork R1V-32B~\cite{peng2025skywork} are multimodal large reasoning models specifically designed for R1-like chain-of-thought capabilities. 
As for 3B models, we compare with LMM-R1-3B~\cite{peng2025lmmr1}, TBAC-VLR1-3B~\cite{Xu2025tbacvlr1}, and VLAA-Thinker-Qwen2.5VL-3B~\cite{vl-thinking2025}.
For models at 7B scale, our comparisons include MM-Eureka-7B~\cite{meng2025mmeureka}, OpenVLThinker-7B~\cite{deng2025openvlthinker}, ThinkLite-VL-7B~\cite{wang2025sota}, and VLAA-Thinker-Qwen2.5VL-7B~\cite{vl-thinking2025}. 
These 3B and 7B models are all trained from Qwen2.5-VL that are enhanced multimodal reasoning through various techniques such as fine-tuning, reinforcement learning, and iterative self-improvement for mathematical reasoning.

\paragraph{Evaluation Benchmarks.}

We evaluate the MLLMs on four prominent multimodal mathematical reasoning benchmarks: MathVision~\cite{wang2025mathvision}, {MathVista}~\cite{lu2023mathvista}, {MathVerse}~\cite{zhang2024mathverse}, and {We-Math}~\cite{qiao2024we}.
These benchmarks offer comprehensive evaluations with diverse problem types, including geometry, charts, and tables, featuring multi-subject math problems and meticulously categorized visual math challenges across various knowledge concepts and granularity levels. We provide our evaluation details in Appendix~\ref{app:bench}.

\subsection{Main Results}\label{sec:benchmark}

In our experiments, we conduct reinforcement learning (RL) using GRPO~\cite{shao2024deepseekmath} with a cold-start strategy.
Specifically, we begin with supervised fine-tuning (SFT) on a distilled dataset generated from Qwen2.5-VL-32B~\cite{bai2025qwen2_5_vl}, which serves as a strong teacher model.
This distilled data is then used to initialize smaller Qwen2.5-VL variants at both the 7B and 3B scales as the cold start stage. Following this initialization, we apply GRPO to further enhance the models’ multimodal reasoning capabilities. The performance of our models is summarized in Table~\ref{tab:main_results}.
Our approach yields consistently strong performance across all four multimodal reasoning benchmarks, demonstrating the effectiveness of our training paradigm.
In particular, our 7B model delivers an overall average improvement of a +6.19 score over the base model (Qwen2.5-VL-7B).
It also achieves state-of-the-art performance among all open-source 7B-scale models, outperforming strong baselines such as MM-Eureka~\cite{meng2025mmeureka}, VLAA-Thinker~\cite{vl-thinking2025}, and ThinkLite-VL~\cite{wang2025sota}. 
Furthermore, our model surpasses many larger models, such as GPT-4o~\cite{hurst2024gpt} and Skywork R1V~\cite{peng2025skywork}. It is also competitive with QvQ-72B~\cite{qwen2024qvq}.
Besides, the 3B variant of our model also sets a new state-of-the-art among open-source 3B multimodal models. It outperforms all existing 3B baselines, including LMM-R1, TBAC-VLR1, and VLAA-Thinker, by significant margins on all four benchmarks.
Notably, our 3B model can achieve competitive results on par with, and in some cases surpassing, larger 7B models such as Qwen2.5-VL-7B and VLAA-Thinker-7B.
In addition, we also calculate the effective rank and the difference ($\Delta \mathrm{eRank}$)~\cite{wei2024large} for models before and after post-training in Table~\ref{tab:diff-erank}.
On a fix-sized model, the value of effective rank usually correlates with the amount of knowledge the model comprehends. After post-training (SFT and RL), extra knowledge is injected into the model, which leads to a consistent increase in effective rank on different benchmarks.
Overall, these results highlight the effectiveness of our cold-start RL pipeline in enhancing reasoning capabilities.
They also demonstrate the potential of scalable training strategies in narrowing the performance gap between small and large multimodal language models.

\begin{table}[t]
    \centering
    \renewcommand{\arraystretch}{1.2}
    \caption{Effective rank and $\Delta \mathrm{eRank}$ of our models before and after post-training on four multimodal reasoning benchmarks.}
    \label{tab:diff-erank}
    \resizebox{0.85\textwidth}{!}{%
    \begin{tabular}{lcccc}
    \toprule
    \textbf{Effective Rank~\cite{wei2024large}} & \textbf{MathVision} & \textbf{MathVerse} & \textbf{MathVista} & \textbf{We-Math}  \\
    \midrule
    Qwen2.5-VL-3B~\cite{bai2025qwen2_5_vl} &74.30 & 66.25 & 66.43 & 60.21 \\ 
    Ours (3B)   &101.68  &  85.91 & 79.59 & 79.96 \\
    \rowcolor{blue!5}
    Difference of Effective Rank ($\Delta \mathrm{eRank}$) & \textcolor{red}{+27.38} & \textcolor{red}{+19.66}  & \textcolor{red}{+13.16} & \textcolor{red}{+19.75} \\
    \midrule
    Qwen2.5-VL-7B~\cite{bai2025qwen2_5_vl} & 73.58 & 62.35 & 63.85 & 60.14 \\
    Ours (7B) & 90.67 & 77.16 & 75.85 & 73.16  \\
    \rowcolor{blue!5}
    Difference of Effective Rank ($\Delta \mathrm{eRank}$)  & \textcolor{red}{+17.09} & \textcolor{red}{+14.81} & \textcolor{red}{+12.00} & \textcolor{red}{+13.02} \\
    \bottomrule
    \end{tabular}%
    } %
    \end{table}

\section{Ablation Studies}

We conduct various ablation studies to explore the interplay between supervised fine-tuning (SFT) and reinforcement learning (RL). The experiments below are conducted based on Qwen2.5-VL-3B.

\subsection{The Effectiveness of Cold Start}

\begin{table}[t]
\centering
\renewcommand{\arraystretch}{1.1}
\caption{Ablation study comparing the effectiveness of different training strategies: Supervised Fine-Tuning (SFT) only, Reinforcement Learning (RL) only, and the combined SFT+RL approach on Qwen2.5-VL models at 3B and 7B scales.}
\label{tab:ablation_results}
\resizebox{0.9\textwidth}{!}{%
\begin{tabular}{>{\raggedright}p{3.0cm}@{\hspace{0.2em}} 
                    *{4}{>{\centering}p{2.5cm}} |
                    >{\centering\arraybackslash}p{2.5cm}}
\toprule
\textbf{Method} & \textbf{MathVision} & \textbf{MathVerse} & \textbf{MathVista} & \textbf{We-Math} & \textbf{Average} \\
\toprule
\rowcolor{gray!10} Qwen2.5-VL-3B~\cite{bai2025qwen2_5_vl} & 19.47 & 33.58 &  56.30 & 50.63 & 40.00\\ 
+ SFT  & 25.20 & {44.39} & 67.60 & 60.63 & 49.46 \\
 + RL  & 25.10 & 39.95 & 66.50 & 63.62 & 48.79 \\
    \rowcolor{blue!5} \textbf{+ SFT and RL}  & {27.04}& 44.03 & {68.20} & {64.08} & \textbf{50.84} \\
\midrule

    \rowcolor{gray!10} Qwen2.5-VL-7B~\cite{bai2025qwen2_5_vl} & 24.87 & 43.83 & 66.30 & 62.87 & 49.47 \\
    + SFT &  28.62 & 46.90 & 71.20 & 67.76 & 53.62 \\
     + RL &  29.80 & 49.29 & {73.50} & 67.82 & 55.10 \\
    \rowcolor{blue!5} \textbf{+ SFT and RL} &  {30.26}  & {48.58} & {73.40} & {70.40} & \textbf{55.66} \\
    
\bottomrule
\end{tabular}%
} %
\end{table}

To better understand the impact of our cold-start strategy, we conduct an ablation study comparing four variants: the base model, RL-only training, SFT-only training (cold start), and the full pipeline combining SFT and RL. The results are presented in Table~\ref{tab:ablation_results}.
In particular, for 3B models, SFT provides a substantial boost in performance over the base model, improving the average score from 40.00 to 49.46. While RL alone also improves performance to 48.79, it is slightly less effective than SFT. 
This suggests that RL without advanced reasoning prior is less effective at guiding the model toward structured problem-solving.
When SFT is followed by RL, the model achieves a new peak of 50.84, demonstrating that the cold-start stage equips the model with strong reasoning patterns, which are further refined by RL.
The 7B models follow a similar trend. Starting from a stronger base (49.47), both SFT (+4.15) and RL (+5.63) lead to notable improvements. The best performance is achieved through the combination of SFT and RL, which yields an average score of 55.66, outperforming either individual component. 

These results collectively highlight that cold-start (SFT) plays a crucial role in unlocking the full potential of reinforcement learning for multimodal reasoning. 
SFT with high-quality CoT data represents a more direct and effective approach to enhancing a model's fundamental reasoning capabilities, rather than relying solely on RL.
When combined, SFT and RL offer a complementary and scalable path to enhance multimodal reasoning.

\begin{table}[t]
\centering
\renewcommand{\arraystretch}{1.15}
\caption{Ablation study on different cold-start strategies for Qwen2.5-VL-3B. We find that the cold start performance correlates well with the final performance.}
\label{tab:ablation_results1}
\resizebox{0.9\textwidth}{!}{%
\begin{tabular}{>{\raggedright}p{4.5cm}@{\hspace{0.2em}} 
                    *{4}{>{\centering}p{1.55cm}} |
                    >{\centering\arraybackslash}p{2.5cm}}
\toprule
\textbf{Strategies} & \textbf{MathVision} & \textbf{MathVerse} & \textbf{MathVista} & \textbf{We-Math} & \textbf{Average} \\
\midrule
\rowcolor{gray!10} None (Base model) & 19.47 & 33.58 & 56.30 & 50.63 & 40.00  \\
\rowcolor{gray!10}\quad + RL  & 25.10 & 39.95 & 66.50 & 63.62 & 48.79 \\
\midrule
Cold Start (32B-Distilled-CoT) & 25.20 & 44.39 & 67.60 & 60.63 & 49.46 \\
\quad + RL  & 27.04 & 44.03 & 68.20 & 64.08 & {50.84} \\[3pt]
Cold Start (7B-Distilled-CoT) & 24.40 & 41.95 & 64.30 & 59.14 & 47.45 \\
\quad + RL & 25.86 & 43.65 & 66.40 & 64.13 & 50.01 \\[3pt]
Cold Start (Caption-CoT) & 22.83 & 40.74 & 62.70 & 58.97 & 46.31 \\
\quad + RL& 25.26 & 42.23 & 65.10 & 63.16 & 48.94\\[3pt]
Cold Start (Reflection-CoT v1) & 23.98 & 40.43 &65.00 & 60.17 & 47.39 \\
\quad + RL& 24.21 & 41.98 & 66.10 & 62.59 & 48.72 \\[3pt]
Cold Start (Self-Critic-CoT) & 23.22 & 40.91 & 61.40 & 57.76 & 45.87 \\
\quad + RL& 25.46 & 41.72 & 66.90 & 64.08& 49.54 \\
\bottomrule
\end{tabular}%
} %
\end{table}

\subsection{Different Types of CoT Data in SFT}

We explore various cold-start strategies beyond direct distillation from a 32B teacher model (Qwen2.5-VL-32B), focusing specifically on the 3B student model. These other strategies are detailed in Section~\ref{sec:cold_start} based on rejection sampling using Qwen2.5-VL-7B.
As shown in Table~\ref{tab:ablation_results1}, the choice of cold-start strategy significantly influences final model performance.
Among all strategies, using CoT data distilled from a strong 32B teacher model yields the best results, achieving an average score of 49.46 after SFT and 50.84 after subsequent RL. 
Direct distillation from Qwen2.5-VL-7B also provides substantial improvements over the base model, but is slightly less effective than the 32B teacher, suggesting that the quality and scale of the teacher model are important for constructing high-quality cold start data.
Other strategies, such as Caption-CoT, Reflection-CoT, and Self-Critic-CoT, also lead to notable gains compared to the base model, but their improvements are generally smaller than those achieved by simple teacher distillation. For example, Caption-CoT and Reflection-CoT (v1) achieve average scores of 46.31 and 47.39 after SFT, respectively, while Self-Critic-CoT yields 45.87. Nevertheless, all these strategies benefit from additional RL, with performance consistently increasing after the RL stage.
In addition, we observe a strong positive correlation between the performance of the cold start stage (SFT) and the final model performance after conducting RL.
This suggests that selecting an effective cold start strategy is crucial for maximizing the overall benefits.
In practice, investing effort in constructing strong SFT data may yield greater performance gains than relying on RL alone to enhance the model's reasoning capabilities.

\begin{figure}[t]
    \centering
    \includegraphics[width=1\textwidth, trim=0cm 0cm 0cm 0cm]{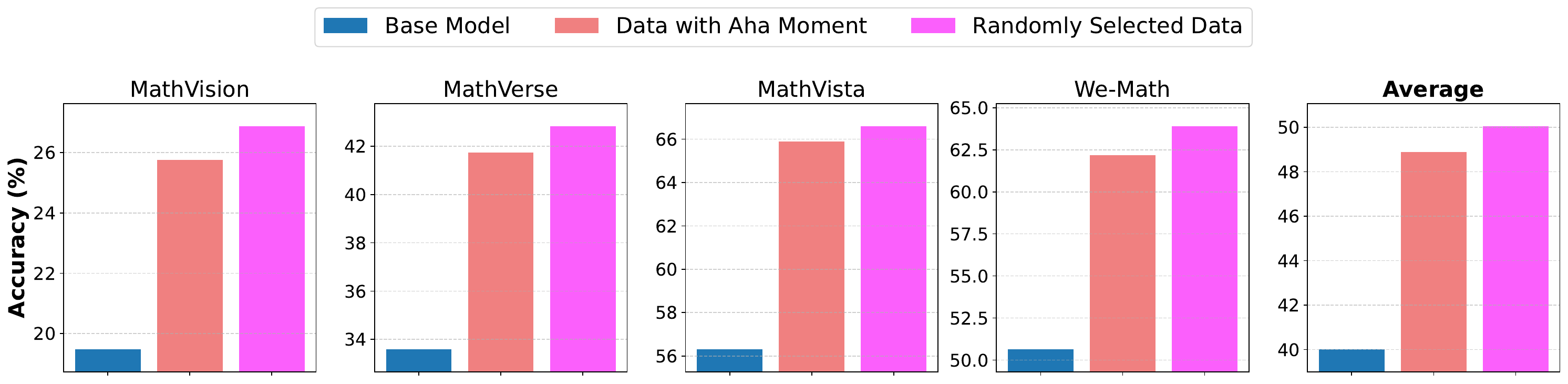}
    \caption{Comparison of model performance when trained on data with ``aha moment'' patterns (Reflection-CoT v2) versus randomly selected 32B-distilled data. Model trained on randomly selected data consistently outperform that trained on ``aha moment'' data, suggesting that these reflective patterns do not necessarily correlate with advanced reasoning capabilities.}
    
    \label{fig:aha}
\end{figure}

\subsection{Revisiting ``Aha Moment''}

As we observe in Section~\ref{sec:observation}, the so-called ``aha moment'' in MLLMs does not necessarily correspond to genuinely useful reasoning patterns. To further investigate this, we conduct a targeted experiment. As described in Section~\ref{sec:cold_start}, we build a dataset by applying rejection sampling to Qwen2.5-VL-32B, retaining only correct responses that exhibit the ``aha moment'' pattern. This yields 10K samples, referred to as Reflection-CoT (v2). For comparison, we also randomly sample 10K instances from the 32B-distilled dataset. 
We then perform SFT on each dataset, followed by GRPO. Results in Figure~\ref{fig:aha} reveal that the model trained on ``aha moment'' data actually underperforms the model trained on randomly selected data. 
This suggests that the presence of the ``aha moment'' in the responses does not necessarily translate into improved reasoning performance.

\begin{table}[t]
    \centering
    \renewcommand{\arraystretch}{1.15}
    \caption{Impact of data quality on model performance. Results show that while rejection sampling (correct data) yields the best performance, models trained on unjudged or even incorrect data still outperform the base model.}
    \label{tab:quality}
    \resizebox{0.9\textwidth}{!}{%
    \begin{tabular}{>{\raggedright}p{4.5cm}@{\hspace{0.2em}} 
                        *{4}{>{\centering}p{1.55cm}} |
                        >{\centering\arraybackslash}p{2.5cm}}
    \toprule
    \textbf{Strategies} & \textbf{MathVision} & \textbf{MathVerse} & \textbf{MathVista} & \textbf{We-Math} & \textbf{Average} \\
    \midrule
    \rowcolor{gray!10}None (Base model) & 19.47 & 33.58 & 56.30 & 50.63 & 40.00 \\
    \rowcolor{gray!10}\quad + RL  & 25.10 & 39.95 & 66.50 & 63.62 & 48.79 \\
    \midrule
    Cold Start (Correct Data) & 25.20 & 44.39 & 67.60 & 60.63 & 49.46 \\
    \quad + RL  & 27.04 & 44.03 & 68.20 & 64.08 & {50.84} \\[3pt]
    Cold Start (Unjudged Data) & 25.59 & 42.08 & 65.80 & 60.06 & 48.38 \\
    \quad + RL& 27.83 & 42.44 & 67.10 & 62.64 & 50.00 \\[3pt]
    Cold Start (Wrong Data) & 24.41 & 37.77 & 63.30 & 55.40 & 45.22 \\
    \quad + RL & 24.44 & 40.38 & 64.70 & 63.51 & 48.26\\
    
    \bottomrule
    \end{tabular}%
    } %
    \end{table}

\subsection{Data Quality in SFT}

We investigate the impact of data quality on model performance by examining different sampling strategies for SFT data collection:
(1) {Rejection Sampling:} Selecting only responses that yield correct answers (our default approach).
(2) {Wrong-Only Sampling:} Deliberately selecting responses that lead to incorrect answers.
(3) {Unjudged Sampling:} Randomly selecting responses without considering correctness.
As shown in Table~\ref{tab:quality}, the highest performance is achieved with rejection sampling, where only correct data is used. Interestingly, models trained on unjudged data—where correctness is not considered—still demonstrate notable improvements in reasoning ability.
Even when trained exclusively on incorrect data, the model outperforms the base model after the cold start phase and continues to benefit from subsequent RL. These results suggest that the structural patterns present in Chain-of-Thought  reasoning can be learned independently of solution correctness, highlighting the importance of reasoning format in addition to answer accuracy.

\section{Conclusion}

In this paper, we demonstrate that ``aha moment'' patterns already exist in MLLMs before RL training but may not necessarily correlate with improved reasoning capabilities. We propose a two-stage approach to enhance multimodal reasoning: first conducting supervised fine-tuning (SFT) with high-quality Chain-of-Thought data to establish a strong foundation, followed by reinforcement learning (RL) to further refine these capabilities. Our extensive experiments show that this combined approach consistently outperforms both SFT-only and RL-only methods, achieving state-of-the-art performance among open-source MLLMs at both 3B and 7B scales. 

\section*{Acknowledgement}
This project is supported by the National Natural Science Foundation of China (No.\ 62406192), Opening Project of the State Key Laboratory of General Artificial Intelligence (No.\ SKLAGI2024OP12), Tencent WeChat Rhino-Bird Focused Research Program, and Doubao LLM Fund.

{\small
\bibliography{reference}
\bibliographystyle{plainnat}}

\clearpage
\appendix
\begin{center}
    \Large \textbf{Appendix}\\[0.5cm]
\end{center}

\section{More Implementation Details}

\subsection{Prompts}\label{app:prompt}

We provide the prompt used to synthesize Self-Critic-CoT~\cite{wen2025thinkpatterns} via rejection sampling.

\begin{tcolorbox}

\textbf{When analyzing any query or task, please follow the structure below:}

\vspace{1em}

\textbf{1. Draft Response:}

Generate an initial response.

\vspace{0.5em}

\textbf{2. Critical Comments:}

Analyze your draft response by considering:

\begin{itemize}
    \item Potential weaknesses or gaps
    \item Logical flaws or inconsistencies
    \item Missing perspectives or alternatives
    \item Areas for improvement
    \item Suggestions for a better version
    \item Steering toward the given answer
\end{itemize}

The critical comments should:

\begin{itemize}
    \item Be specific and actionable
    \item Reference particular parts of the draft
    \item Suggest concrete improvements
    \item Consider different angles or approaches
    \item Guide towards a more comprehensive solution
\end{itemize}

\vspace{1em}

\textbf{Output Format:}

\begin{itemize}
    \item \textbf{Draft Response:} \\
    Your initial complete response to the instruction. 
    \item \textbf{Critical Comments:} \\
    Your analysis of the draft response, highlighting areas for improvement and
    suggesting specific enhancements.
    \item \textbf{Final Answer:} \\
    Put your final answer within \verb|\boxed{}|.
\end{itemize}
\end{tcolorbox}

\subsection{Rejection Sampling}\label{app:rej_samp}

To construct the distillation datasets from Qwen2.5-VL-7B and Qwen2.5-VL-32B, we employ a rejection sampling technique~\cite{tong2024dart}. We try at most 24 times to obtain one correct response. We observe that this approach results in approximately the same amount of data (both 52K samples) for Qwen2.5-VL-7B and Qwen2.5-VL-32B.

\subsection{Benchmarks}\label{app:bench}

We provide some details about the benchmarks we use to evaluate the models' reasoning ability.
{MathVision}~\cite{wang2025mathvision} is a challenging benchmark containing 3040 mathematical problems with visual contexts from real-world math competitions across 12 grades. It covers 16 subjects over 5 difficulty levels, including specialized topics like Analytic Geometry, Combinatorial Geometry, and Topology.

\begin{itemize}[itemsep=0pt,topsep=0pt,leftmargin=0.5cm]
    \item \textbf{MathVision}~\cite{wang2025mathvision} is a challenging benchmark containing 3040 mathematical problems with visual contexts from real-world math competitions across 12 grades. It covers 16 subjects over 5 difficulty levels, including specialized topics like Analytic Geometry, Combinatorial Geometry, and Topology.

    \item \textbf{MathVista}~\cite{lu2023mathvista} is a comprehensive benchmark for evaluating mathematical reasoning in visual contexts. It contains 1000 questions featuring diverse problem types including geometry, charts, and tables.
    
    \item \textbf{MathVerse}~\cite{zhang2024mathverse} is an all-around visual math benchmark designed for an equitable and in-depth evaluation of MLLMs. The test set contains 3940 multi-subject math problems with diagrams from publicly available sources, focusing on Plane Geometry and Solid Geometry.
    
    \item \textbf{We-Math}~\cite{qiao2024we} meticulously collect and categorize 1740 visual math problems in the test set, spanning 67 hierarchical knowledge concepts and 5 layers of knowledge granularity. 
\end{itemize}

For all benchmarks, we prompt the models to place their final answers within a designated box format. We then employ Qwen2.5-32B-Instruct~\cite{yang2024qwen2} to evaluate answer correctness by comparing the extracted responses with ground truth answers, which often contain complex mathematical expressions. Note that our reported benchmark scores may differ from those in the original papers due to variations in evaluation protocols.

\subsection{Training Details}

We conduct post-training using Qwen2.5-VL-3B and Qwen2.5-VL-7B~\cite{bai2025qwen2_5_vl}.
In particular, we employ ms-swift~\cite{zhao2024swiftascalablelightweightinfrastructure} framework for the cold start training (SFT), using 3 epochs with a learning rate of \(1 \times 10^{-5}\).
Besides, we adopt the EasyR1~\cite{zheng2025easyr1} framework for GRPO training. We mainly follow the default hyperparameters in EasyR1.
We set the training episodes to 2 with a learning rate of \(1 \times 10^{-6}\).
During GRPO training, we adopt accuracy reward, which assigns rewards solely based on the correctness of the generated response: a correct final answer receives a reward of +1, while an incorrect one receives a reward of 0. 

\section{Compute Resources}

We conduct our experiments using NVIDIA H800-80G and A800-40G GPUs. The experimental time using 8
A800 for training Qwen2.5-VL-7B using GRPO is around 12 hours.

\section{Limitations}

While our study provides comprehensive insights into the interplay between supervised fine-tuning (SFT) and reinforcement learning (RL) for multimodal reasoning, several limitations remain.

Firstly, we primarily evaluate models at the 3B and 7B parameter scales due to limited computation resources. The generalizability of our findings to larger models (e.g., 30B, 70B, or beyond) remains to be validated.
Our approach adopts GRPO~\cite{shao2024deepseekmath} as the RL optimization algorithm, which is currectly most popular. We do not systematically compare alternative RL methods, which we leave as future work.
In addition, the identification of ``aha moment'' patterns is based on keyword matching and heuristic rules, which may introduce subjectivity or incompleteness.

\section{Case Studies}

We provide several detailed case studies of our observation and curated CoT datasets as follows.

\subsection{``Aha Moment'' in MLLM}\label{app:aha_case}

In the first case, the correct answer should be 130°, but the model fails to arrive at this solution. The associated reflection pattern exhibits hallucinations and proves ineffective. In contrast, in the second case, the model successfully answers the question, and the reflection pattern contributes meaningfully to the reasoning process.
\begin{tcolorbox}[floatplacement=t, 
  title=\textbf{The First Case.},
  fonttitle=\small,
  colbacktitle=gray!20,
  breakable,
  coltitle=black]
\tiny

\begin{center}
    \includegraphics[width=0.2\textwidth, trim=0cm 0cm 0cm 0cm]{figs/36.jpg}
    \label{fig:app1}
\end{center}

\textbf{Question: }

In the given parallelogram ABCD, if the sum of angle B and angle D measures 100°, what is the measure of angle A?

\textbf{Model's response:}

To solve this problem, we need to use the properties of a parallelogram. The properties are:

\begin{enumerate}
  \item Opposite sides are parallel.
  \item Opposite angles are equal.
  \item Consecutive angles are supplementary (they add up to \(180^\circ\)).
\end{enumerate}

Given:
The sum of angle \(B\) and angle \(D\) is \(100^\circ\).

Since \(AD \parallel BC\) and \(AB \parallel CD\), angles \(B\) and \(D\) are consecutive angles. Therefore, they are supplementary, meaning they add up to \(180^\circ\).

So, we can write:
\[
\text{angle } B + \text{angle } D = 180^\circ
\]

Given:
\[
\text{angle } B + \text{angle } D = 100^\circ
\]

This implies that the given sum (\(100^\circ\)) is incorrect for the sum of angles \(B\) and \(D\) in a parallelogram, which should be \(180^\circ\). However, let's proceed with the given information.

Since angle \(B\) and angle \(D\) are consecutive angles, we can find the measure of angle \(A\) by using the fact that opposite angles are equal in a parallelogram. Therefore, angle \(A\) is equal to angle \(C\).

Now, assume:
\[
\text{angle } B = x,\quad \text{angle } D = 180^\circ - x
\]

Given:
\[
x + (180^\circ - x) = 100^\circ
\Rightarrow 180^\circ = 100^\circ
\]

\textcolor{red}{This is incorrect, so we need to \textbf{re-evaluate} the problem. The correct interpretation should be that the sum of angle \(B\) and angle \(D\) is \(180^\circ\), not \(100^\circ\).}

Therefore, the measure of angle \(A\) is:
\[
\boxed{140^\circ}
\]
\end{tcolorbox}

\begin{tcolorbox}[floatplacement=t, 
  title=\textbf{The Second Case.},
  fonttitle=\small,
  colbacktitle=gray!20,
  breakable,
  coltitle=black]
\begin{center}
    \includegraphics[width=0.2\textwidth, trim=0cm 0cm 0cm 0cm]{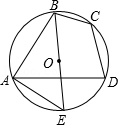}
\end{center}

\tiny
\textbf{Question: }

In the given diagram, quadrilateral ABCD is inscribed in circle O. The diameter BE is part of circle O and AE is connected to it. If the measure of angle BCD is twice the measure of angle BAD, then what is the degree measure of angle DAE?

\textbf{Model's response:}

To solve the problem, we need to use some properties of cyclic quadrilaterals and the relationships between angles in a circle.

\begin{enumerate}
    \item Identify the given information:
    \begin{itemize}
        \item Quadrilateral \(ABCD\) is inscribed in circle \(O\).
        \item Diameter \(BE\) of circle \(O\) means that \(BE\) is a straight line passing through the center of the circle.
        \item \(AE\) is connected to \(B\) and \(E\).
        \item \(\angle BCD = 2 \times \angle BAD\).
    \end{itemize}
    
    \item Use the property of cyclic quadrilaterals:
    
    In a cyclic quadrilateral, the sum of the opposite angles is \(180^\circ\). Therefore,
    \[
    \angle BAD + \angle BCD = 180^\circ.
    \]
    
    \item Express \(\angle BCD\) in terms of \(\angle BAD\):
    
    Given \(\angle BCD = 2 \times \angle BAD\), substitute this into the equation:
    \[
    \angle BAD + 2 \times \angle BAD = 180^\circ,
    \]
    \[
    3 \times \angle BAD = 180^\circ,
    \]
    \[
    \angle BAD = 60^\circ,
    \]
    \[
    \angle BCD = 2 \times 60^\circ = 120^\circ.
    \]
    
    \item Determine \(\angle ADE\):
    
    Since \(BE\) is a diameter, \(\angle ADE\) is a right angle because the angle subtended by a diameter in a semicircle is \(90^\circ\):
    \[
    \angle ADE = 90^\circ.
    \]
    
    \item Determine \(\angle DAE\):
    
    \(\angle DAE\) is part of triangle \(ADE\), where \(\angle ADE = 90^\circ\). The sum of the angles in triangle \(ADE\) is \(180^\circ\):
    \[
    \angle DAE + \angle ADE + \angle AED = 180^\circ,
    \]
    \[
    \angle DAE + 90^\circ + \angle AED = 180^\circ,
    \]
    \[
    \angle DAE + \angle AED = 90^\circ.
    \]
    
    \item Determine \(\angle AED\):
    
    \(\angle AED\) is the same as \(\angle BCD\) because they subtend the same arc \(CD\):
    \[
    \angle AED = 120^\circ.
    \]
    
    \item Solve for \(\angle DAE\):
    
    Substitute \(\angle AED = 120^\circ\) into the equation:
    \[
    \angle DAE + 120^\circ = 90^\circ,
    \]
    \[
    \angle DAE = -30^\circ.
    \]
    
    \textcolor{red}{Since \(\angle DAE\) cannot be negative, we must \textbf{re-evaluate} the problem. The correct interpretation is that \(\angle AED = 60^\circ\) and not \(120^\circ\), because \(\angle AED\) does not subtend the same arc as \(\angle BCD\). Instead, the triangle should be analyzed differently, considering interior angles.}

    Thus, the correct value is:
    \[
    \boxed{30^\circ}.
    \]
\end{enumerate}
\end{tcolorbox}

\subsection{Distilled Data from Different Models}\label{app:distill}

The main difference between the Qwen2.5-VL-7B distilled CoT and the Qwen2.5-VL-32B distilled CoT lies in their level of detail and organization. The Qwen2.5-VL-32B distilled CoT is more comprehensive and better structured, offering richer, more informative reasoning steps compared to the more concise and less detailed Qwen2.5-VL-7B distilled CoT.

\begin{tcolorbox}[floatplacement=t, 
  title=\textbf{The Distilled CoT Data from Qwen2.5-VL-7B and Qwen2.5-VL-32B.},
  fonttitle=\small,
  colbacktitle=gray!20,
  breakable,
  coltitle=black]
\begin{center}
    \includegraphics[width=0.2\textwidth, trim=0cm 0cm 0cm 0cm]{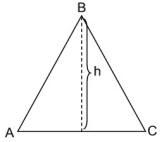}
\end{center}

\tiny
\textbf{Question: }

In the accompanying diagram, ABC is an equilateral triangle with a perimeter of 30. What is the length of altitude h of this triangle?

Choices: $5\sqrt2, 5\sqrt3, 10\sqrt2, 10\sqrt3$

\rule{\linewidth}{0.5pt}

\textbf{Qwen2.5-VL-7B's response:}

To find the length of the altitude \( h \) of an equilateral triangle with a perimeter of 30, we can follow these steps:

\begin{enumerate}
    \item Calculate the side length of the equilateral triangle:
    \begin{itemize}
        \item The perimeter of an equilateral triangle is the sum of the lengths of its three sides.
        \item Given the perimeter is 30, we can find the length of one side by dividing the perimeter by 3:
        \[
        \text{Side length} = \frac{30}{3} = 10
        \]
    \end{itemize}

    \item Use the properties of an equilateral triangle:
    \begin{itemize}
        \item In an equilateral triangle, the altitude also bisects the base, creating two 30-60-90 right triangles.
        \item In a 30-60-90 triangle, the ratio of the sides opposite the \(30^\circ\), \(60^\circ\), and \(90^\circ\) angles is \(1 : \sqrt{3} : 2\).
    \end{itemize}

    \item Determine the length of the altitude \( h \):
    \begin{itemize}
        \item The altitude \( h \) is the side opposite the \(60^\circ\) angle in the right triangle.
        \item The side opposite the \(30^\circ\) angle (which is half the base) is 5 (since the base is 10 and it is bisected).
        \item Using the ratio \( 1 : \sqrt{3} : 2 \), the altitude \( h \) is:
        \[
        h = 5 \sqrt{3}
        \]
    \end{itemize}
\end{enumerate}

Therefore, the length of the altitude \( h \) is:

\[
\boxed{5\sqrt{3}}
\]

\rule{\linewidth}{0.5pt}

\textbf{Qwen2.5-VL-32B's response:}

Step-by-Step Solution:

1. Understand the Problem:
\begin{itemize}
    \item We are given an equilateral triangle \( \triangle ABC \) with a perimeter of 30.
    \item We need to find the length of the altitude \( h \) of this triangle.
\end{itemize}

2. Find the Side Length of the Triangle:
\begin{itemize}
    \item An equilateral triangle has all three sides of equal length.
    \item Let the side length of the triangle be \( s \).
    \item The perimeter of the triangle is given by:
    $\text{Perimeter} = s + s + s = 3s$
    \item We are given that the perimeter is 30:
    $3s = 30$

    \item Solving for \( s \):
    $
    s = \frac{30}{3} = 10
    $
    \item Therefore, the side length of the triangle is \( s = 10 \).
\end{itemize}

3. Use the Formula for the Altitude of an Equilateral Triangle:
\begin{itemize}
    \item The altitude \( h \) of an equilateral triangle with side length \( s \) is given by the formula:
    \[
    h = \frac{\sqrt{3}}{2} s
    \]
    \item Substitute \( s = 10 \) into the formula:
    \[
    h = \frac{\sqrt{3}}{2} \cdot 10
    \]
    \item Simplify the expression:
    \[
    h = 5\sqrt{3}
    \]
\end{itemize}

4. Verify the Result:
\begin{itemize}
    \item The formula for the altitude of an equilateral triangle is derived from the Pythagorean theorem.
    \item In an equilateral triangle, the altitude bisects the base, creating two 30-60-90 right triangles.
    \item In a 30-60-90 triangle, the sides are in the ratio \( 1 : \sqrt{3} : 2 \).
    \item The altitude corresponds to the side opposite the 60-degree angle, which is \( \frac{\sqrt{3}}{2} \) times the hypotenuse (the side of the equilateral triangle).
    \item This confirms that the formula \( h = \frac{\sqrt{3}}{2} s \) is correct.
\end{itemize}

5. Select the Correct Answer:
\begin{itemize}
    \item The length of the altitude \( h \) is \( 5\sqrt{3} \).
\end{itemize}

\begin{center}
\[
\boxed{5\sqrt{3}}
\]
\end{center}
\end{tcolorbox}

\end{document}